\documentclass[11pt]{article}

\usepackage[utf8]{inputenc}
\usepackage[T1]{fontenc}
\usepackage{hyperref}
\usepackage{graphicx}
\usepackage{booktabs}
\usepackage{amsmath}
\usepackage{array}
\usepackage{fvextra}
\usepackage{tikz}
\usetikzlibrary{arrows.meta,positioning}

\DefineVerbatimEnvironment{Code}{Verbatim}{
  breaklines=true,
  breakanywhere=true,
  fontsize=\small
}

\title{Graph-Based Agentic AI with LangGraph: Workflow Pathways for Long-Running Stateful Business Processes}

\author{
Daniel Pearson\\
Dhillon School of Business, University of Lethbridge\\
Lethbridge, Alberta, Canada\\[0.75em]
Sidney Shapiro\\
Dhillon School of Business, University of Lethbridge\\
Lethbridge, Alberta, Canada\\[0.75em]
Emiliano Sebastian Gonzalez Venegas\\
Universidad de Guadalajara\\
Guadalajara, Jalisco, Mexico\\[0.75em]
Sanad Al-Khatib\\
Al Hussein Technical University\\
Amman, Jordan\\[0.75em]
Aurora Pinzón Arzola\\
Universidad de Guanajuato\\
León, Guanajuato, Mexico
}

\date{}

\begin{document}

\maketitle

\begin{abstract}
This paper is a practitioner guide to graph-based workflow pathways for long-running, stateful, multi-step generative AI systems in business processes. Rather than treating LangGraph, a low-level orchestration framework for stateful agents, as a model-quality benchmark target, we present three executable recipes---SQL analytics with repair loops, agentic retrieval-augmented generation with evidence gating, and human-in-the-loop policy review with interrupt and checkpoint recovery---to show how typed state, conditional routing, deterministic tools, retries, interrupts, checkpoints, and traces fit together. LangGraph is positioned by workflow-complexity fit, not as a universal default: simpler ReAct-style or plain SDK loops may be better for basic tool use, schema-first tools for structured extraction and validation, and DSPy when prompt or program optimization is the main goal. Each recipe explains when LangGraph is worth the extra structure and which implementation patterns make routes, pauses, and audit trails explicit product behavior rather than hidden prompt logic.
\end{abstract}

\section{Introduction}

Consider a policy-review assistant used inside a business process. A user asks whether a high-risk employment action is allowed. A simple LLM call can draft an answer, but it cannot by itself enforce the process: gather policy passages, score risk, pause for human review, persist the pending state, resume after approval, apply reviewer feedback, and leave a decision record. LangGraph is useful when those steps are not incidental implementation details but part of the product contract.

LangGraph is a low-level orchestration framework for building, managing, and deploying long-running stateful agents and workflows~\cite{langgraph2024}. It is not itself a language model, a prompt library, or an end-to-end product agent. Instead, it supplies the control-plane primitives---typed shared state, graph nodes and edges, conditional routing, durable checkpoints, and interrupt/resume---that turn multi-step LLM applications into inspectable process graphs. In that sense it sits below higher-level agent packages: developers keep ownership of tools, prompts, and domain logic, while LangGraph makes the workflow structure explicit.

What LangGraph does in practice is make durable execution and governance first-class. Workflows can persist through failures and pauses and resume from the last checkpoint; humans can inspect or modify state at interrupt points; typed working state carries intermediate artifacts across nodes; and route history supports debugging and audit rather than leaving control flow only in model prose. Separate long-lived memory stores are supported by the framework but are outside the three recipes exercised here. Those properties matter when tool use, retries, branching, persistent state, and approval gates are part of the product contract rather than incidental implementation detail.

In the policy-review example above, that contract becomes a state graph: a typed state object, nodes such as \texttt{draft\_decision} and \texttt{risk\_score}, conditional edges to \texttt{finalize\_decision} or \texttt{interrupt\_for\_review}, and a checkpointer so an interrupted thread can resume later. Use LangGraph when the application needs durable state across steps, visible branch decisions, repair paths after failure, human review gates, or audit trails. Do not start with LangGraph when the task is a single prompt, one tool call, simple structured extraction, or primarily a prompt-optimization problem. In those cases, a plain SDK or ReAct-style loop, a schema-first agent, or DSPy is usually clearer (Table~\ref{tab:alternatives-fit}). This paper is a pathway guide: it explains how to assemble common business-process patterns as executable LangGraph workflows and when the graph abstraction is worth using.

The paper deliberately avoids treating LangGraph as a model-quality contest. The useful question is not whether a graph makes an LLM smarter; it is whether a graph makes a workflow easier to inspect, repair, pause, resume, and govern. We therefore focus on recipes for three recurring pathways: SQL analytics, agentic retrieval-augmented generation (RAG), and human-in-the-loop (HITL) policy review.

What this guide provides is a decision guide that organizes when to use LangGraph versus a plain SDK/ReAct loop, a schema-first agent, or DSPy (Table~\ref{tab:alternatives-fit}); three concrete LangGraph recipes that illustrate SQL repair, evidence-aware RAG, and checkpointed HITL review; code patterns that package typed state, node boundaries, conditional routing, retries, interrupts, and checkpoints; and trade-off guidance that shows where graph orchestration adds value and where it adds unnecessary complexity.

\section{When LangGraph Fits}
\label{sec:when-to-use}

The first design decision is whether the problem is actually a workflow problem. LangGraph is usually justified when at least one of the following holds: the process can pause and resume later, especially for human review or long-running work; the next step depends on explicit state such as risk level, validation status, evidence quality, or retry count; failure should route to a repair path rather than simply returning an error; the team needs a trace of which route was taken and why; or multiple tools or model calls must share a durable state object.

LangGraph is usually unnecessary when the task is a short linear call: prompt, tool call, structured response, done. Avoid it as a first choice when the workflow has no branch and no durable state, human approval happens outside the system and never resumes inside the graph, the team only needs final inputs and outputs in logs rather than route history, or a single retry wrapper around one model call would solve the failure mode. In those cases, graph structure adds concepts---state schemas, builders, route functions, checkpointers---without clearer product behavior. The recipe question should therefore be: \emph{what state must survive, what decisions must be explicit, and what failures need a route?}

\subsection{Decision flow}

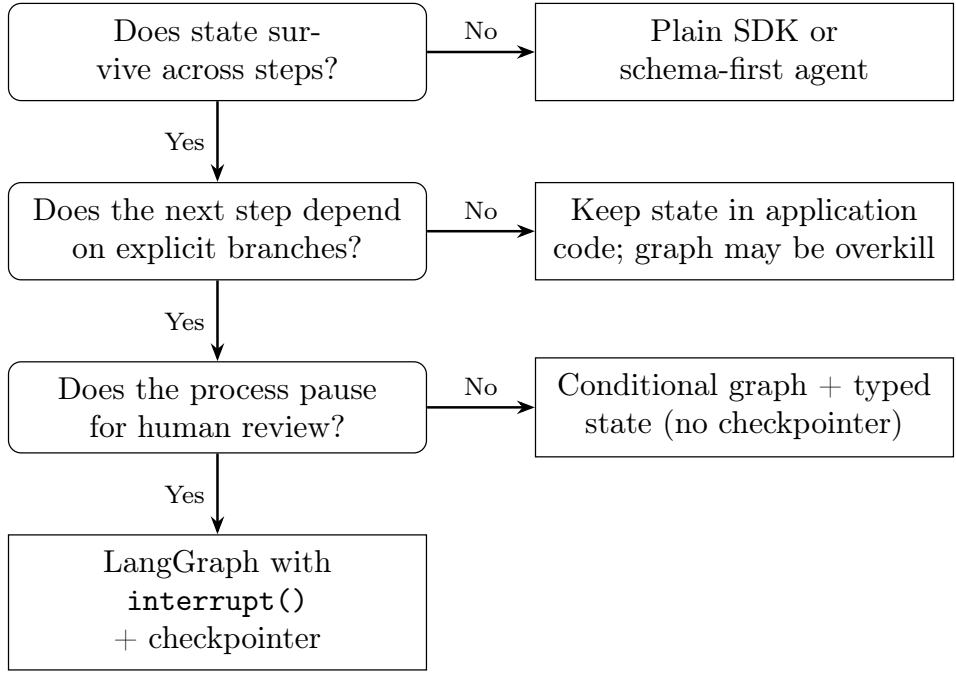
\begin{center}
\resizebox{\linewidth}{!}{%
\begin{tikzpicture}[
  decision/.style={
    draw,
    rounded corners,
    align=center,
    text width=0.34\linewidth,
    inner sep=5pt,
    font=\small
  },
  outcome/.style={
    draw,
    align=center,
    text width=0.34\linewidth,
    inner sep=5pt,
    font=\small
  },
  arrow/.style={-{Stealth[length=2mm]}, thick},
  node distance=9mm and 12mm
]
\node[decision] (state) {Does state survive across steps?};
\node[outcome, right=of state] (plain) {Plain SDK or schema-first agent};
\node[decision, below=of state] (branch) {Does the next step depend on explicit branches?};
\node[outcome, right=of branch] (appstate) {Keep state in application code; graph may be overkill};
\node[decision, below=of branch] (review) {Does the process pause for human review?};
\node[outcome, right=of review] (nocp) {Conditional graph + typed state (no checkpointer)};
\node[outcome, below=of review] (checkpoint) {LangGraph with \texttt{interrupt()} + checkpointer};

\draw[arrow] (state) -- node[above, font=\scriptsize] {No} (plain);
\draw[arrow] (state) -- node[left, font=\scriptsize] {Yes} (branch);
\draw[arrow] (branch) -- node[above, font=\scriptsize] {No} (appstate);
\draw[arrow] (branch) -- node[left, font=\scriptsize] {Yes} (review);
\draw[arrow] (review) -- node[above, font=\scriptsize] {No} (nocp);
\draw[arrow] (review) -- node[left, font=\scriptsize] {Yes} (checkpoint);
\end{tikzpicture}
}
\end{center}

The figure above is intentionally minimal. Repair routes after validation or retrieval failure and audit requirements for route history also favor a conditional graph even when human review is absent; Table~\ref{tab:langgraph-decision} covers those cases.

Table~\ref{tab:langgraph-decision} makes the decision concrete by pairing common workflow requirements with the LangGraph mechanism that addresses them and the simpler case where a graph is probably unnecessary.

\begin{table}[htbp]
\centering
\caption{Decision table for when to use LangGraph.}
\label{tab:langgraph-decision}
\small
\begin{tabular}{@{}p{0.28\linewidth}p{0.36\linewidth}p{0.26\linewidth}@{}}
\toprule
Use LangGraph when & Why LangGraph helps & Use something simpler when \\
\midrule
The workflow must pause and resume, especially for human review or background work
  & Checkpointers and interrupts preserve state across a review, delay, failure, or process restart
  & Approval happens entirely outside the application and no in-graph resume is needed \\
The next step depends on explicit state such as risk, evidence quality, validation status, or retry count
  & Conditional edges make route decisions inspectable instead of hiding them in prompt text or nested application logic
  & The process is a short linear call with no meaningful branch \\
Failures require repair routes rather than a single error response
  & Retry budgets, error state, and repair nodes give failures a controlled path back into the workflow
  & A simple retry wrapper around one model or tool call solves the failure mode \\
The team needs auditability for which branch ran and why
  & Node boundaries and shared state create a traceable decision record for debugging, review, or compliance
  & Final inputs and outputs are enough for the operational log \\
Multiple model/tool calls must coordinate through shared durable state
  & A typed state object keeps intermediate artifacts, route decisions, and final outputs explicit across nodes
  & The task is primarily structured extraction, schema validation, or prompt/program optimization \\
\bottomrule
\end{tabular}
\end{table}

\subsection{Alternatives at a glance}

LangGraph should not be treated as a universal replacement for every LLM application framework. Table~\ref{tab:alternatives-fit} summarizes where simpler approaches may fit better. A plain SDK or ReAct-style loop is often sufficient for basic tool use when durable checkpoints and human approval are unnecessary~\cite{yao2023react}. Schema-first systems are often a better fit for structured extraction and validation-heavy tasks~\cite{geng2025jsonschemabench,dagdelen2024structured,pydanticai2026}. DSPy is often a better fit when the primary goal is metric-guided prompt or program optimization~\cite{khattab2024dspy,opsahlong2024mipro}. LangGraph's APIs are most relevant when the application needs explicit state management, branching control, recovery from failure, human intervention, and durable workflow state~\cite{langgraph2024}; agent and memory surveys motivate why those needs recur~\cite{wang2023agentsurvey,zhang2024memorysurvey}.

\begin{table}[htbp]
\centering
\caption{Alternatives to LangGraph by problem class. This is a selection guide, not an empirical ranking of frameworks.}
\label{tab:alternatives-fit}
\footnotesize
\setlength{\tabcolsep}{4pt}
\begin{tabular}{@{}p{0.22\linewidth}p{0.36\linewidth}p{0.32\linewidth}@{}}
\toprule
Alternative & Better case than LangGraph & Research support \\
\midrule
Plain SDK / simple ReAct loop
  & Simple tool use or question answering without durable checkpoints, graph state, or human approval
  & Lightweight reason--act loops can work with prompting alone~\cite{yao2023react}. \\
Schema-first tools (e.g.\ PydanticAI-style validation)
  & Structured extraction, JSON output, form filling, and validation-heavy tasks
  & Schema compliance is a major task class~\cite{geng2025jsonschemabench}; scientific extraction produces useful structured records~\cite{dagdelen2024structured}; schema-first agents emphasize typed validation~\cite{pydanticai2026}. \\
DSPy
  & When the main problem is prompt/program optimization rather than workflow orchestration
  & Metric-guided compilation of LM pipelines~\cite{khattab2024dspy}; MIPRO optimizes instructions and demonstrations for multi-stage programs~\cite{opsahlong2024mipro}. \\
LangGraph
  & Long-running or branching workflows with typed state, checkpoints, retries, and human review
  & Graph APIs for stateful multi-actor apps~\cite{langgraph2024}; agent surveys motivate planning, tools, and structure~\cite{wang2023agentsurvey}; memory surveys motivate persistent state~\cite{zhang2024memorysurvey}; reflection architectures motivate iterative recovery~\cite{shinn2023reflexion}. \\
\bottomrule
\end{tabular}
\end{table}

\section{Graph Anatomy and Repository Layout}
\label{sec:anatomy}

LangGraph models long-running stateful workflows as \texttt{StateGraph} objects with nodes, edges, conditional routing, subgraphs, and optional checkpointers for durable execution~\cite{langgraph2024}. A developer moves control flow out of informal application code and into an inspectable graph: state is passed between nodes, edge functions decide the next step, and checkpoints can persist execution across failures, pauses, and human review.

\subsection{Core recipe parts}

A LangGraph recipe has four core parts, plus an optional design concern for auditability. \textbf{Typed state} is the durable record passed through the workflow, including user input, intermediate artifacts, route decisions, retry counts, and final outputs. \textbf{Nodes} are small functions that do one step, such as retrieving documents, validating SQL, scoring risk, or requesting review. \textbf{Conditional edges} are routing functions that turn state into the next step, such as \texttt{execute}, \texttt{repair}, \texttt{retrieve\_again}, \texttt{interrupt}, or \texttt{finalize}. \textbf{Persistence} is a checkpointer when the workflow must survive a pause, process restart, approval delay, or background job boundary. A useful design concern, but not a separate runtime primitive claimed as exercised here, is keeping enough structured state and route history to reconstruct what happened after the fact.

Table~\ref{tab:workflows} summarizes how the three recipes in this paper instantiate those parts. Section~\ref{sec:code} gives the code shape for branches, retry loops, interrupt points, and checkpoint/resume behavior.

\begin{table}[htbp]
\centering
\caption{Workflow pathway recipe summary.}
\label{tab:workflows}
\small
\begin{tabular}{@{}p{0.20\linewidth}p{0.36\linewidth}p{0.33\linewidth}@{}}
\toprule
Pathway & Nodes & Why graph it? \\
\midrule
SQL analytics & Schema lookup, SQL generation, validation, execution, summary & Validation and execution errors need explicit repair routes \\
Agentic RAG & Question analysis, retrieval, document grading, answer generation, citation verification & Evidence quality should decide whether to answer, retrieve again, or clarify \\
HITL policy review & Draft, risk score, interrupt, feedback, finalize & High-risk decisions need durable human review and resumable state \\
\bottomrule
\end{tabular}
\end{table}

\subsection{Use cases and pattern coverage}

LangGraph is used in production-oriented LLM systems where workflows span multiple steps, tools, and approval gates. Table~\ref{tab:usecases} summarizes representative use cases, the graph patterns they imply, and whether each pattern is exercised by the executable pathway examples. Rows marked \emph{No} or \emph{Partial} name adjacent patterns---customer support, multi-agent research, and session-spanning business-process management---that this guide references but does not implement as full recipes (Section~\ref{sec:scope}).

\begin{table}[htbp]
\centering
\caption{Representative LangGraph use cases and recipe coverage.}
\label{tab:usecases}
\small
\begin{tabular}{@{}>{\raggedright\arraybackslash}p{2.2cm}>{\raggedright\arraybackslash}p{3.8cm}>{\raggedright\arraybackslash}p{2.8cm}c@{}}
\toprule
Use case & Typical graph pattern & Pathway example & Exercised \\
\midrule
SQL analytics & Generate--validate--execute--repair loop & SQL analytics & Yes \\
Agentic RAG & Conditional retrieve--grade--answer & Agentic RAG & Yes \\
Policy / compliance review & Draft--risk score--interrupt--finalize & HITL policy review & Yes \\
Customer support & Classify--retrieve--draft--escalate & --- & No \\
Multi-agent research & Supervisor--worker subgraphs & --- & No \\
Long-running BPM & Durable checkpoints across sessions & Partial (HITL) & Partial \\
\bottomrule
\end{tabular}
\end{table}

\textbf{SQL analytics agents} map naturally to cyclic graphs: schema lookup, generation, validation, execution, and repair loops~\cite{yu2018spider}. \textbf{Agentic RAG} requires conditional retrieval, evidence grading, and refusal or retry when support is weak~\cite{lewis2020rag}; ReAct-style loops motivate interleaving retrieval and generation without requiring a durable graph~\cite{yao2023react}. \textbf{Human-in-the-loop policy review} requires durable interrupts, reviewer feedback, and resumable execution---a primary motivation for checkpointed graph state~\cite{langgraph2024}.

Design patterns recurring across these cases include: (1) shared typed state passed between nodes; (2) conditional edges for retry and escalation; (3) tool nodes with deterministic backends; and (4) interrupt nodes for external review. The examples implement patterns (1)--(3) in all three workflows and pattern (4) in the HITL workflow.

\subsection{Repository layout}

The repository keeps each pathway small enough to inspect. Each workflow has a \texttt{graph.py} file for LangGraph structure and a \texttt{logic.py} file for domain logic. This separation is important: the graph should express routing and state transitions; the logic file should contain SQL validation, retrieval scoring, policy checks, and other ordinary application behavior.

The examples can run with mock model responses for repeatable local development or with live providers for integration testing. That execution harness is secondary to the paper. Its purpose is to make the recipes runnable, not to rank frameworks.

The most important implementation constraint is to keep node boundaries stable. Good node boundaries correspond to product events such as schema lookup, SQL validation, retrieval, document grading, risk scoring, and interruption for review. If a step would be useful in an audit log, a retry rule, or a dashboard, it is a candidate graph node.

\section{Three Pathway Recipes}
\label{sec:recipes}

This section gives the recipes directly. Table~\ref{tab:recipe-cards} is a recipe map, not a performance table: it summarizes the state, routes, and stop conditions that define each pathway.

\begin{table}[htbp]
\centering
\caption{Recipe cards for the three LangGraph pathways.}
\label{tab:recipe-cards}
\footnotesize
\setlength{\tabcolsep}{3pt}
\begin{tabular}{@{}p{0.17\linewidth}p{0.24\linewidth}p{0.26\linewidth}p{0.24\linewidth}@{}}
\toprule
Pathway & Key state fields & Critical routes & LangGraph justified when \\
\midrule
SQL analytics & \texttt{sql}, \texttt{error}, \texttt{attempts}, \texttt{rows} & Validation: retry or execute; execution: retry or summarize & Repair loops and auditability matter \\
Agentic RAG & \texttt{evidence\_grade}, \texttt{citations}, \texttt{retries} & Grade: retry or generate; verify: retry or finalize & Evidence gating is product behavior \\
HITL review & \texttt{risk\_level}, \texttt{approved}, \texttt{decision\_record} & Risk: interrupt or finalize & Durable human gate is required \\
\bottomrule
\end{tabular}
\end{table}

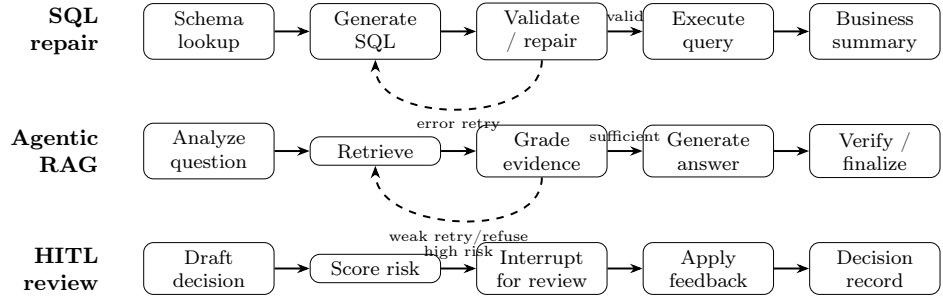
\begin{figure}[htbp]
\centering
\resizebox{\linewidth}{!}{%
\begin{tikzpicture}[
  boxnode/.style={
    draw,
    rounded corners,
    align=center,
    inner sep=3pt,
    font=\scriptsize,
    text width=0.13\linewidth
  },
  labelnode/.style={
    align=right,
    font=\footnotesize\bfseries,
    text width=0.11\linewidth
  },
  arrow/.style={-{Stealth[length=1.8mm]}, thick},
  node distance=5mm and 5mm
]
\node[labelnode] (sqlLabel) {SQL repair};
\node[boxnode, right=of sqlLabel] (sqlSchema) {Schema lookup};
\node[boxnode, right=of sqlSchema] (sqlGenerate) {Generate SQL};
\node[boxnode, right=of sqlGenerate] (sqlValidate) {Validate / repair};
\node[boxnode, right=of sqlValidate] (sqlExecute) {Execute query};
\node[boxnode, right=of sqlExecute] (sqlSummary) {Business summary};

\node[labelnode, below=8mm of sqlLabel] (ragLabel) {Agentic RAG};
\node[boxnode, right=of ragLabel] (ragAnalyze) {Analyze question};
\node[boxnode, right=of ragAnalyze] (ragRetrieve) {Retrieve};
\node[boxnode, right=of ragRetrieve] (ragGrade) {Grade evidence};
\node[boxnode, right=of ragGrade] (ragAnswer) {Generate answer};
\node[boxnode, right=of ragAnswer] (ragVerify) {Verify / finalize};

\node[labelnode, below=8mm of ragLabel] (hitlLabel) {HITL review};
\node[boxnode, right=of hitlLabel] (hitlDraft) {Draft decision};
\node[boxnode, right=of hitlDraft] (hitlRisk) {Score risk};
\node[boxnode, right=of hitlRisk] (hitlReview) {Interrupt for review};
\node[boxnode, right=of hitlReview] (hitlFeedback) {Apply feedback};
\node[boxnode, right=of hitlFeedback] (hitlRecord) {Decision record};

\draw[arrow] (sqlSchema) -- (sqlGenerate);
\draw[arrow] (sqlGenerate) -- (sqlValidate);
\draw[arrow] (sqlValidate) -- node[above, font=\tiny] {valid} (sqlExecute);
\draw[arrow] (sqlExecute) -- (sqlSummary);
\draw[arrow, dashed] (sqlValidate.south) to[out=-90,in=-90] node[below, font=\tiny] {error retry} (sqlGenerate.south);

\draw[arrow] (ragAnalyze) -- (ragRetrieve);
\draw[arrow] (ragRetrieve) -- (ragGrade);
\draw[arrow] (ragGrade) -- node[above, font=\tiny] {sufficient} (ragAnswer);
\draw[arrow] (ragAnswer) -- (ragVerify);
\draw[arrow, dashed] (ragGrade.south) to[out=-90,in=-90] node[below, font=\tiny] {weak retry/refuse} (ragRetrieve.south);

\draw[arrow] (hitlDraft) -- (hitlRisk);
\draw[arrow] (hitlRisk) -- node[above, font=\tiny] {high risk} (hitlReview);
\draw[arrow] (hitlReview) -- (hitlFeedback);
\draw[arrow] (hitlFeedback) -- (hitlRecord);
\end{tikzpicture}
}
\caption{High-level route sketches for the three LangGraph pathway recipes. The HITL row omits the low-risk finalize shortcut for space; see Figure~\ref{fig:hitl-interrupt}. The graph abstraction matters where validation errors, weak evidence, or human-review risk should determine the next step explicitly.}
\label{fig:pathway-recipes}
\end{figure}

Figure~\ref{fig:pathway-recipes} gives the same recipes as route sketches: each row shows the ordinary forward path and the decision point that makes the workflow worth graphing.

\subsection{SQL analytics repair loop}

\textbf{Problem fit.} Use this recipe when natural-language database queries need a controlled repair path: validation errors and execution failures should route back to generation until a retry budget is exhausted or a valid result can be summarized.

\textbf{State schema.} \texttt{SQLState} carries the question and schema; generated SQL, errors, and attempt count; result columns and rows; and the final answer and status.

\textbf{Node sequence.} Schema lookup leads to SQL generation and validation. A conditional route then selects query execution, retry, business summary, or failure.

\textbf{Routing rules.} After validation, \texttt{route\_after\_validation} returns \texttt{execute} when \texttt{error} is empty, \texttt{retry} when attempts remain, otherwise \texttt{fail}. After execution, \texttt{route\_after\_execution} returns \texttt{summarize}, \texttt{retry}, or \texttt{fail} using the same pattern.

\textbf{Worked example (\texttt{sql\_tool\_failure}).} Input: ``What is total revenue by region?'' The graph looks up schema, generates SQL, and hits a validation or execution error on the first attempt. State records the error and increments \texttt{attempts}. The conditional edge routes to \texttt{sql\_generation} again with the error message as repair context. On a successful execution, the graph routes to \texttt{business\_summary} and terminates with \texttt{status=completed}.

\begin{Code}
schema_lookup -> sql_generation -> sql_validation
  |-- error + attempts left --> sql_generation (retry)
  |-- error + budget exhausted --> fail
  |-- valid SQL --> query_execution
        |-- execution error + attempts left --> sql_generation
        |-- success --> business_summary -> END
\end{Code}

\textbf{When not to use.} Prefer a plain SDK chain when the query is never repaired, never audited, and failures can return a single error response (Table~\ref{tab:alternatives-fit}). Prefer a schema-first agent when the bottleneck is typed SQL or result validation without a multi-step recovery graph.

\subsection{Agentic RAG evidence loop}

\textbf{Problem fit.} Use this recipe when retrieval quality should determine whether the workflow answers, retrieves again, asks for clarification, or refuses. Weak evidence should be a route, not a hidden prompt instruction.

\textbf{State schema.} \texttt{RAGState} carries the question and retrieval decision; retrieved documents and their evidence grade; the answer and citations; and retry and status fields.

\textbf{Node sequence.} \texttt{question\_analysis} $\rightarrow$ retrieve or generate $\rightarrow$ \texttt{grade\_documents} $\rightarrow$ \texttt{generate\_answer} $\rightarrow$ \texttt{verify\_citations} $\rightarrow$ \texttt{finalize} or \texttt{retry\_or\_clarify}.

\textbf{Routing rules.} \texttt{route\_after\_grade} sends weak evidence back through retrieval while retries remain. Once that budget is exhausted, generation produces an explicit insufficiency response, and citation verification leads to a failed final status rather than a supported answer. \texttt{route\_after\_verify} likewise retries failed citation checks only while its retry budget allows.

\textbf{Worked example (\texttt{rag\_weak\_evidence}).} Input: ``What is the policy on xyz123?'' Retrieval returns low-quality documents, \texttt{evidence\_grade} is \texttt{weak}, and the graph routes to \texttt{retry\_or\_clarify}. After retries are exhausted, the workflow terminates with \texttt{status=failed} rather than fabricating a supported answer. That terminal failure is correct pathway behavior, not a system crash.

\begin{Code}
question_analysis -> retrieve -> grade_documents
  |-- weak evidence + retries left --> retry_or_clarify -> retrieve
  |-- otherwise --> generate_answer -> verify_citations
        |-- bad citations + retries left --> retry_or_clarify
        |-- supported answer --> finalize -> END
        |-- unsupported --> finalize (failed) -> END
\end{Code}

\textbf{When not to use.} Prefer a plain SDK or ReAct-style retrieve-then-answer loop when evidence gating, citation verification, and retry semantics are not product requirements (Table~\ref{tab:alternatives-fit}). Schema-first tools help if the output must be a typed record, but they do not by themselves enforce weak-evidence routes.

\subsection{HITL policy review with interrupt and checkpoint}

\textbf{Problem fit.} Use this recipe when a process must pause for a human decision and resume later with the same state. High-risk policy questions are the clearest case: the workflow should stop at a review boundary, wait for approval, and continue only after reviewer input is recorded.

\textbf{State schema.} \texttt{ReviewState} carries the draft answer, risk level, and policy passages; approval and reviewer feedback; and the final answer, decision record, and status.

\textbf{Node sequence.} \texttt{draft\_decision} $\rightarrow$ \texttt{risk\_score} $\rightarrow$ conditional route $\rightarrow$ \texttt{interrupt\_for\_review} or \texttt{finalize\_decision}. When review is required, the graph pauses at \texttt{interrupt()}, resumes with reviewer input, applies feedback, and finalizes a decision record.

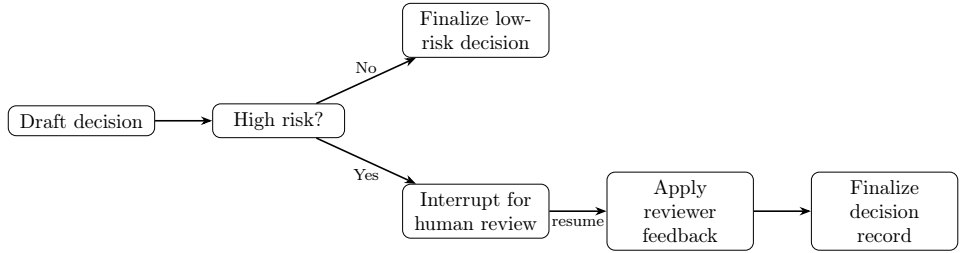
\begin{figure}[htbp]
\centering
\resizebox{\linewidth}{!}{%
\begin{tikzpicture}[
  boxnode/.style={
    draw,
    rounded corners,
    align=center,
    inner sep=4pt,
    font=\small,
    text width=0.18\linewidth
  },
  decision/.style={
    draw,
    rounded corners,
    align=center,
    inner sep=4pt,
    font=\small,
    text width=0.16\linewidth
  },
  arrow/.style={-{Stealth[length=2mm]}, thick},
  node distance=8mm and 10mm
]
\node[boxnode] (draft) {Draft decision};
\node[decision, right=of draft] (risk) {High risk?};
\node[boxnode, above right=of risk] (finalLow) {Finalize low-risk decision};
\node[boxnode, below right=of risk] (interrupt) {Interrupt for human review};
\node[boxnode, right=of interrupt] (feedback) {Apply reviewer feedback};
\node[boxnode, right=of feedback] (record) {Finalize decision record};

\draw[arrow] (draft) -- (risk);
\draw[arrow] (risk) -- node[above, font=\scriptsize] {No} (finalLow);
\draw[arrow] (risk) -- node[below, font=\scriptsize] {Yes} (interrupt);
\draw[arrow] (interrupt) -- node[below, font=\scriptsize] {resume} (feedback);
\draw[arrow] (feedback) -- (record);
\end{tikzpicture}
}
\caption{HITL interrupt/resume route. The graph pauses only on high-risk state, exposes the draft and evidence to a reviewer, and resumes from the checkpoint after reviewer feedback is supplied.}
\label{fig:hitl-interrupt}
\end{figure}

\textbf{Routing rules.} \texttt{route\_after\_risk} routes to \texttt{interrupt\_for\_review} when \texttt{risk\_level} is \texttt{high}, \texttt{force\_human\_review} is set, or the task requires escalation.

\textbf{Worked example (\texttt{hitl\_high\_risk}).} Input: ``Can we terminate an employee immediately for expense fraud?'' The graph drafts an answer, scores risk as high, and routes to \texttt{interrupt\_for\_review}. The interrupt payload exposes \texttt{draft\_answer}, \texttt{risk\_level}, and \texttt{policy\_passages} to the reviewer. Execution pauses until \texttt{Command(resume=\{...\})} supplies approval and feedback. The graph then applies feedback, finalizes the decision, and writes a structured \texttt{decision\_record}.

\begin{Code}
draft_decision -> risk_score
  |-- low risk --> finalize_decision -> END
  |-- high risk --> interrupt_for_review (pause)
        resume(reviewer_input) -> apply_feedback -> finalize_decision -> END
\end{Code}

\textbf{Design warnings.} Removing the checkpointer breaks durable pause/resume: a high-risk case cannot complete after interruption because there is no persisted thread state. Removing the review gate can leave \texttt{status=completed} while the decision record no longer matches what a reviewer would have approved. Both variants show why checkpointing and explicit review routes are part of the recipe, not optional polish.

\textbf{When not to use.} If human approval is handled entirely outside the workflow and the process never resumes inside the graph, an ordinary queue or ticket system is usually simpler than a checkpointed interrupt (Table~\ref{tab:alternatives-fit}). Schema-first validation and DSPy-style prompt optimization do not substitute for durable pause/resume: interrupt plus checkpointer is the justified LangGraph cell for this pathway.

\section{Python Workflow Examples}
\label{sec:code}

This section gives compact Python examples for the three recurring pathways. The examples are shortened from the repository implementations, but they keep the important LangGraph structure: typed state, nodes, conditional edges, and termination routes.

\subsection{SQL analytics: validate, repair, execute}

\textbf{Context.} A business user asks a natural-language analytics question. The workflow should generate SQL, validate it, execute it, and repair failures rather than returning the first broken query.

\begin{Code}
class SQLState(TypedDict):
    question: str
    schema: str
    sql: str
    error: str
    attempts: int
    rows: list[list]
    final_answer: str
    status: str

builder = StateGraph(SQLState)
builder.add_node("schema_lookup", schema_lookup)
builder.add_node("sql_generation", sql_generation)
builder.add_node("sql_validation", sql_validation)
builder.add_node("query_execution", query_execution)
builder.add_node("business_summary", business_summary)
builder.add_node("fail", fail)

builder.add_edge(START, "schema_lookup")
builder.add_edge("schema_lookup", "sql_generation")
builder.add_edge("sql_generation", "sql_validation")
builder.add_conditional_edges(
    "sql_validation",
    route_after_validation,
    {"execute": "query_execution",
     "retry": "sql_generation",
     "fail": "fail"},
)
builder.add_conditional_edges(
    "query_execution",
    route_after_execution,
    {"summarize": "business_summary",
     "retry": "sql_generation",
     "fail": "fail"},
)
builder.add_edge("business_summary", END)
builder.add_edge("fail", END)
graph = builder.compile()
\end{Code}

\textbf{Discussion.} This is a good LangGraph fit because validation and execution errors are not ordinary exceptions; they are workflow states. The graph makes the repair path explicit: a validation error routes to \texttt{sql\_generation}, a successful query routes to \texttt{business\_summary}, and exhausted retries route to \texttt{fail}. If the application only needs one generated query and can return errors directly, this graph is unnecessary ceremony.

\subsection{Agentic RAG: retrieve, grade, verify citations}

\textbf{Context.} A RAG assistant should not answer when evidence is weak or citations fail verification. Retrieval quality should control the route through the workflow.

\begin{Code}
class RAGState(TypedDict):
    question: str
    needs_retrieval: bool
    retrieved_docs: list[dict]
    evidence_grade: str
    answer: str
    citations: list[str]
    retries: int
    status: str

builder = StateGraph(RAGState)
builder.add_node("question_analysis", question_analysis)
builder.add_node("retrieve", retrieve)
builder.add_node("grade_documents", grade_documents)
builder.add_node("generate_answer", generate_answer)
builder.add_node("verify_citations", verify_citations)
builder.add_node("retry_or_clarify", retry_or_clarify)
builder.add_node("finalize", finalize)

builder.add_edge(START, "question_analysis")
builder.add_conditional_edges(
    "question_analysis",
    route_after_analysis,
    {"retrieve": "retrieve", "generate": "generate_answer"},
)
builder.add_edge("retrieve", "grade_documents")
builder.add_conditional_edges(
    "grade_documents",
    route_after_grade,
    {"generate": "generate_answer", "retry": "retry_or_clarify"},
)
builder.add_edge("retry_or_clarify", "retrieve")
builder.add_edge("generate_answer", "verify_citations")
builder.add_conditional_edges(
    "verify_citations",
    route_after_verify,
    {"finalize": "finalize", "retry": "retry_or_clarify"},
)
builder.add_edge("finalize", END)
graph = builder.compile()
\end{Code}

\textbf{Discussion.} The important design move is treating evidence quality as state. Weak evidence or bad citations should not be buried in a prompt instruction such as ``only answer if supported.'' They should become route labels that can be tested, traced, and changed. LangGraph is useful when the product must distinguish \texttt{finalize}, \texttt{retry}, and \texttt{clarify}; a simpler RAG chain is enough when every query follows one retrieve-generate path.

\subsection{HITL policy review: interrupt, resume, checkpoint}

\textbf{Context.} A policy workflow drafts an answer, scores risk, and pauses for human review when the decision is high risk. The process must resume later with the same state and record the reviewer input.

\begin{Code}
class ReviewState(TypedDict):
    question: str
    policy_passages: list[dict]
    draft_answer: str
    risk_level: str
    approved: bool
    reviewer_feedback: str
    final_answer: str
    decision_record: str
    status: str

def interrupt_for_review(state: ReviewState) -> dict:
    reviewer = interrupt({
        "draft_answer": state["draft_answer"],
        "risk_level": state["risk_level"],
        "policy_passages": state["policy_passages"],
    })
    return {
        "approved": reviewer.get("approved", False),
        "reviewer_feedback": reviewer.get("feedback", ""),
    }

builder = StateGraph(ReviewState)
builder.add_node("draft_decision", draft_decision)
builder.add_node("risk_score", risk_score)
builder.add_node("interrupt_for_review", interrupt_for_review)
builder.add_node("apply_feedback", apply_feedback)
builder.add_node("finalize_decision", finalize_decision)
builder.add_edge(START, "draft_decision")
builder.add_edge("draft_decision", "risk_score")
builder.add_conditional_edges(
    "risk_score",
    route_after_risk,
    {"interrupt_for_review": "interrupt_for_review",
     "finalize_decision": "finalize_decision"},
)
builder.add_edge("interrupt_for_review", "apply_feedback")
builder.add_edge("apply_feedback", "finalize_decision")
builder.add_edge("finalize_decision", END)
graph = builder.compile(checkpointer=InMemorySaver())
\end{Code}

\begin{Code}
config = {"configurable": {"thread_id": case["task_id"]}}
graph.invoke(initial_state, config)  # pauses at interrupt()
graph.invoke(Command(resume=reviewer_input), config)
\end{Code}

\textbf{Discussion.} This is the clearest case for LangGraph. The interrupt is not just a callback; it is a durable workflow boundary. The checkpointer preserves the thread so the graph can resume after review without reconstructing state from logs or a ticketing system. \texttt{InMemorySaver()} is appropriate for local tests; production HITL should use a durable checkpointer (SQLite, Postgres, or equivalent) so \texttt{thread\_id} state survives process restarts and delayed review. If approval happens outside the application and no graph state needs to resume, an ordinary queue is probably simpler.

\subsection{Route functions should stay small}

Across all three examples, route functions should read state and return small labels.

\begin{Code}
def route_after_validation(state):
    if not state["error"]:
        return "execute"
    if state["attempts"] <= retry_limit():
        return "retry"
    return "fail"

def route_after_grade(state):
    if state["evidence_grade"] == "weak":
        if state["retries"] < retry_limit():
            return "retry"
        # Generate an insufficiency response; verification then fails closed.
    return "generate"

def route_after_risk(state):
    if state["risk_level"] == "high" or state.get("force_human_review"):
        return "interrupt_for_review"
    return "finalize_decision"
\end{Code}

Keeping routes small is what makes the graph inspectable. The model may draft content, but application state decides whether the workflow retries, escalates, fails closed, or finalizes.

\section{Trade-offs and Testing}
\label{sec:tradeoffs}

\textbf{Complexity.} LangGraph adds concepts: state schemas, graph builders, route functions, checkpointers, and invocation configuration. That overhead is worthwhile only when those concepts map to real product needs.

\textbf{Explicit routing.} The main benefit is not better prose from the model. It is the ability to say which branch ran, why it ran, and what state caused the next step. This matters for repair loops, compliance review, and operational debugging.

\textbf{Durable pauses.} HITL workflows are the clearest case for LangGraph. An interrupt with a checkpointer gives a concrete place to stop, hand state to a reviewer, and resume without rebuilding the process from scratch.

\textbf{Checkpointers.} The HITL example compiles with \texttt{InMemorySaver()} for local tests. That suffices when the graph, reviewer, and resume call run in one process during development. Production HITL and background jobs need a durable checkpointer backed by SQLite, Postgres, or another store so \texttt{thread\_id} state survives restarts and delayed review. The SQL and RAG recipes omit checkpointers because they terminate in one session; add one only when the same pathway must pause or resume across process boundaries.

\textbf{Audit traces.} Auditable workflows should persist route-relevant fields in typed state---for example \texttt{attempts}, \texttt{evidence\_grade}, \texttt{risk\_level}, \texttt{status}, and structured outputs such as \texttt{decision\_record}---rather than relying on model prose alone. Node boundaries then double as trace events: each transition updates fields a test or log can inspect without re-running the LLM.

\textbf{Retry and escalation choices.} Immediate retries are appropriate when the failure message is useful input for the next attempt. Clarification routes are better than hallucinated final answers when evidence is weak. Escalation routes should fire when risk, policy, or confidence crosses a threshold. Fail-closed terminations are preferable to unsupported answers when the workflow cannot justify a response.

\textbf{Testing.} Recipe tests should assert route behavior and state transitions, not answer quality alone. For example, a contract test might assert \texttt{route\_after\_validation(state) == "retry"} when \texttt{error} is set and \texttt{attempts} remain, or that a resumed HITL thread preserves \texttt{draft\_answer} while updating \texttt{reviewer\_feedback}. Broader expectations: bad SQL should route to \texttt{retry} until the retry budget is exhausted; weak evidence should not reach \texttt{finalize} with a supported answer; high-risk policy decisions should reach \texttt{interrupt\_for\_review}; and resumed HITL cases should preserve thread state and produce a \texttt{decision\_record}. The repository uses small fixture cases for these contract checks. They verify workflow behavior without turning the paper into a benchmark report.

\textbf{Avoiding overuse.} If a workflow has no branch, no durable state, no retry semantics, no human pause, and no audit requirement, LangGraph is probably not the right first abstraction.

\section{Related Work}
\label{sec:related}

\textbf{Graph-based agent orchestration.} LangGraph provides a concrete API for representing LLM workflows as stateful graphs with checkpointing and human interrupts~\cite{langgraph2024}. The broader pattern---durable control flow with shared state---is not exclusive to one library. Multi-agent conversation frameworks such as AutoGen~\cite{wu2023autogen} and role-based crew orchestrators such as CrewAI~\cite{crewai2024} address overlapping agent-coordination needs, often with less explicit checkpointing. Durable workflow engines such as Temporal~\cite{temporal2024} provide long-running execution, retries, and human tasks that can wrap LLM steps at the infrastructure layer. This paper uses LangGraph as the concrete programming surface for recipes while treating the underlying pattern as transferable. Prior work on ReAct-style reasoning and acting~\cite{yao2023react} and Toolformer-style tool use~\cite{schick2023toolformer} shows that lightweight tool-using loops can succeed without durable graph orchestration; persistence and branching then remain implicit in application code. Agent and memory surveys motivate planning, tools, evaluation, and persistent state as recurring design needs for long-horizon agents~\cite{wang2023agentsurvey,zhang2024memorysurvey}. Reflexion and generative-agent architectures likewise motivate memory, reflection, and long-horizon state as first-class concerns~\cite{shinn2023reflexion,park2023generativeagents}. This paper treats those ideas as engineering patterns: when control flow is itself a product artifact, a graph makes the routes inspectable.

\textbf{Retrieval-augmented generation.} RAG pipelines combine retrieval, generation, and verification stages~\cite{lewis2020rag}. Agentic variants add conditional retrieval and self-correction loops. Our agentic RAG recipe follows that pattern by grading evidence and refusing to finalize unsupported answers. Public retrieval benchmarks such as BEIR are useful references for the problem class~\cite{thakur2021beir}, but this paper focuses on the routing recipe rather than benchmark scores.

\textbf{Text-to-SQL agents.} Semantic parsing benchmarks such as WikiSQL, Spider, and BIRD illustrate natural-language database querying under different schema and realism assumptions~\cite{zhong2017seq2sql,yu2018spider,li2023bird}. Our SQL pathway uses the same problem shape---generate, validate, execute, repair---as a reusable business-analytics recipe rather than as a leaderboard entry.

\textbf{Policy and legal reasoning.} LegalBench provides a public multi-task benchmark for legal reasoning, including privacy-policy question answering~\cite{guha2023legalbench}. We use that domain as motivation for the HITL policy-review recipe: escalation, evidence support, and decision-record completeness are workflow requirements even when the system is not offering legal advice.

\textbf{Programmatic LLM workflows.} The alternatives in Table~\ref{tab:alternatives-fit} map to distinct problem classes rather than a single winner. Plain ReAct-style loops are often sufficient for simple tool use~\cite{yao2023react}. Schema-first stacks are often a better fit when the dominant need is typed or constrained output: JSON Schema benchmarks measure schema compliance~\cite{geng2025jsonschemabench}, scientific extraction work shows structured records such as JSON objects from text~\cite{dagdelen2024structured}, and PydanticAI-style agents emphasize validation~\cite{pydanticai2026}. Agno-style wrappers similarly emphasize role workflows~\cite{agno2026}. DSPy targets metric-guided compilation of LM programs~\cite{khattab2024dspy}, with MIPRO optimizing instructions and demonstrations across multi-stage pipelines~\cite{opsahlong2024mipro}. Those approaches are complementary to LangGraph: they are often the better first choice for linear, schema-centric, or optimization-centric tasks. LangGraph's APIs for durable state, review gates, and auditable routing~\cite{langgraph2024} are most relevant when those orchestration properties are product requirements; agent and memory surveys motivate why such properties recur in long-horizon settings~\cite{wang2023agentsurvey,zhang2024memorysurvey}.

\section{Practical Implications}
\label{sec:implications}

For practitioners designing a workflow pathway, the guidance is a decision tree rather than a single recommendation (Table~\ref{tab:alternatives-fit}):

\begin{enumerate}
  \item \textbf{Start with workflow requirements.} Write down the state that must survive, the route decisions that matter, and the failure paths users expect.
  \item \textbf{Use a plain SDK first for linear work.} If the process is prompt, maybe one tool, then response, keep it simple.
  \item \textbf{Prefer schema-first when the bottleneck is typed or validated output.} Structured extraction, JSON forms, and record validation often need schemas more than graphs.
  \item \textbf{Prefer DSPy when the bottleneck is optimizing prompts or programs across examples.} Metric-guided instruction and demonstration search is a different problem from durable orchestration.
  \item \textbf{Move to LangGraph when routes become product behavior.} Validation failure, weak evidence, review escalation, and retry exhaustion are routes worth making explicit.
  \item \textbf{Add checkpointing only when pause/resume is real.} Checkpointers are most valuable for HITL review, background jobs, and long-running processes.
  \item \textbf{Treat traces as part of the interface.} The graph should make it possible to reconstruct what happened without reading model prose.
\end{enumerate}

Organizations already standardized on other agent layers can still use LangGraph selectively for subgraphs that require durable HITL state, visible routing, or compliance traces.

\section{Scope and Availability}
\label{sec:scope}

This paper is a recipe and design guide, not an empirical benchmark. It does not estimate production accuracy, latency, cost, or model quality. The examples are intended to show where graph orchestration changes the engineering shape of an application.

\subsection{Availability}

The reference implementation needed to run the three recipes is included with this arXiv version as ancillary material under \path{anc/}. It contains the LangGraph package source, prompts, small fixture datasets, and three focused workflow tests. It excludes live credentials and the legacy benchmark-result archive because neither is needed to exercise the pathway examples. Appendix~\ref{sec:runnable} gives the local commands.

\subsection{Limitations}

Three executable recipes cover SQL repair, agentic RAG, and HITL policy review (Table~\ref{tab:usecases}). They deliberately omit multi-agent supervisor--worker subgraphs, customer-support escalation playbooks, and cross-session memory stores; those appear in Table~\ref{tab:usecases} as adjacent use cases rather than demonstrated pathways. The code examples use typed in-graph state and, for HITL, an in-memory checkpointer suitable for local development; production deployments should swap in a durable checkpointer (for example SQLite or Postgres) when pause/resume must survive process restarts. We do not evaluate streaming UX, observability backends, deployment topology, or operational cost. Schema-first validation and DSPy-style prompt optimization can complement LangGraph inside individual nodes; combining stacks is noted in Section~\ref{sec:implications} but not shown as a fourth recipe.

\section{Conclusion}

We presented three executable LangGraph workflow pathways for long-running business processes: SQL analytics with repair loops, agentic RAG with evidence loops, and HITL policy review with interrupt/checkpoint recovery. The point is not that LangGraph improves model quality. The point is that graph orchestration gives developers explicit places to store state, route failures, pause for review, resume after interruption, and explain what happened. LangGraph is not a universal replacement for every LLM application stack. Choose by problem class: a plain SDK or ReAct-style loop for simple tool use, schema-first tools for structured extraction and validation, DSPy when prompt or program optimization is the main goal, and LangGraph when stateful branching, recovery, human review, and long-running coordination are product requirements.

\appendix
\section{Runnable Examples}
\label{sec:runnable}

Minimal local setup after downloading and extracting the complete arXiv source package (mock mode; no provider keys):
\begin{Code}
cd anc
python3.10 -m venv .venv
source .venv/bin/activate
pip install -e ".[dev]"
export LANGGRAPH_STUDY_MODE=mock
python -m pytest tests/test_recipes.py -q
\end{Code}

The same fixture IDs used in the pathway recipes are available as case tasks (for example \texttt{sql\_tool\_failure}, \texttt{rag\_weak\_evidence}, and \texttt{hitl\_high\_risk} under \texttt{benchmarks/datasets/}). Happy-path tests above exercise the three graphs; failure and review cases assert repair, refuse/retry, and interrupt/resume routes rather than model quality.

The three LangGraph pathway implementations are intentionally small and inspectable: SQL analytics in \path{src/langgraph_study/workflows/sql_analytics/graph.py}, agentic RAG in \path{src/langgraph_study/workflows/agentic_rag/graph.py}, and HITL policy review in \path{src/langgraph_study/workflows/hitl_policy_review/graph.py}.

Each pathway follows the same shape:
\begin{Code}
builder = StateGraph(TypedState)
builder.add_node("node_name", node_fn)
builder.add_edge(START, "first_node")
builder.add_conditional_edges("route_node", route_fn, route_map)
builder.add_edge("terminal_node", END)
graph = builder.compile(checkpointer=...)
\end{Code}

Set \texttt{LANGGRAPH\_STUDY\_MODE=live} (mock/live run mode) and provider API keys to exercise the same recipes with live model providers.

\bibliographystyle{plain}
\bibliography{references}

\end{document}